\documentclass{article}

\usepackage{PRIMEarxiv}
\usepackage[utf8]{inputenc}
\usepackage[T1]{fontenc}
\usepackage{tabularx}
\usepackage{cite}
\usepackage{amsmath,amssymb,amsfonts}
\usepackage{graphicx}
\usepackage{booktabs}
\usepackage{multirow}
\usepackage{array}
\usepackage{hyperref}
\usepackage{url}
\hypersetup{hidelinks}
\usepackage{textcomp}
\usepackage{microtype}
\graphicspath{{media/}}
\newcommand{\IEEEPARstart}[2]{#1#2}
\begin{document}

\title{PA-TCNet: Pathology-Aware Temporal Calibration with Physiology-Guided Target Refinement for Cross-Subject Motor Imagery EEG Decoding in Stroke Patients}

\author{
\parbox{0.92\textwidth}{\centering
Xiangkai Wang$^{1}$, Yun Zhao$^{2,*}$, Dongyi He$^{1,3}$, Qingling Xia$^{1}$, Gen Li$^{1,4}$,\\
Nizhuan Wang$^{3,*}$, Ningxiao Peng$^{5}$, and Bin Jiang$^{1,*}$\\
$^{1}$School of Artificial Intelligence, Chongqing University of Technology, Chongqing 401135, China;\\
Chongqing Key Laboratory of Embodied Intelligence Perception and Autonomous Learning\\
for Humanoid Robots; Key Laboratory of Advanced Equipment Intelligence\\
of the Chongqing Education Commission, Chongqing 401135, China\\
$^{2}$School of Smart Health, Chongqing Polytechnic University of Electronic Technology, Chongqing 401131, China\\
$^{3}$Department of Language Science and Technology, The Hong Kong Polytechnic University,\\
Hung Hom, Hong Kong SAR, China\\
$^{4}$School of Pharmacy and Bioengineering, Chongqing University of Technology, Chongqing 400054, China\\
$^{5}$School of Computer Science and Engineering, Chongqing University of Technology, Chongqing 401135, China\\
\texttt{kevinwang@stu.cqut.edu.cn; zhaoyun@cqcet.edu.cn; hedongyi6438@gmail.com}\\
\texttt{qingling@cqut.edu.cn; ligen1990@cqut.edu.cn; wangnizhuan1120@gmail.com}\\
\texttt{xxx\_saik@163.com; jb20200132@cqut.edu.cn}\\
\small $^{*}$Corresponding authors: Yun Zhao (\texttt{zhaoyun@cqcet.edu.cn}),\\
\small Nizhuan Wang (\texttt{wangnizhuan1120@gmail.com}), and\\
\small Bin Jiang (\texttt{jb20200132@cqut.edu.cn})
}
}

\maketitle

\begin{abstract}
Stroke patient cross-subject electroencephalography (EEG) decoding of motor imagery (MI) brain-computer interface (BCI) is essential for motor rehabilitation, yet lesion-related abnormal temporal dynamics and pronounced inter-patient heterogeneity often undermine generalization. Existing adaptation methods are easily misled by pathological slow-wave activity and unstable target-domain pseudo-labels. To address this challenge, we propose PA-TCNet, a pathology-aware temporal calibration framework with physiology-guided target refinement for stroke motor imagery decoding. PA-TCNet integrates two coordinated components. The Pathology-aware Rhythmic State Mamba (PRSM) module decomposes EEG spatiotemporal features into slowly varying rhythmic context and fast transient perturbations, injecting the fused pathological context into selective state propagation to more effectively capture abnormal temporal dynamics. The Physiology-Guided Target Calibration (PGTC) module constructs source-domain sensorimotor region-of-interest templates, imposing physiological consistency constraints and dynamically refining target-domain pseudo-labels, thereby improving adaptation reliability. Leave-one-subject-out experiments on two independent stroke EEG datasets, XW-Stroke and 2019-Stroke, yielded mean accuracies of 66.56\% and 72.75\%, respectively, outperforming state-of-the-art baselines. These results indicate that jointly modeling pathological temporal dynamics and physiology-constrained pseudo-supervision can provide more robust cross-subject initialization for personalized post-stroke MI-BCI rehabilitation. The implemented code is available at \url{https://github.com/wxk1224/PA-TCNet}.

\end{abstract}


\section{Introduction}
\label{sec:intro}

\IEEEPARstart{S}{troke} is one of the leading neurological disorders causing long-term motor dysfunction, and it is commonly characterized by impaired upper-limb motor function, reduced coordination, and prolonged recovery duration\cite{Feigin2022,Wilson2016}. Motor imagery (MI) electroencephalography (EEG) decoding based on brain-computer interfaces (BCIs) provides important technical support for recognizing patients' motor intentions and facilitating subsequent rehabilitation training\cite{JIANG2026109680, 10.3389/fneur.2026.1672882 }. Owing to its non-invasive nature, high temporal resolution, portability, and relatively low cost, EEG has become one of the most important signal carriers for MI decoding in intelligent rehabilitation engineering\cite{Abiri2019}.

Among related tasks, cross-subject decoding is a key prerequisite for clinical translation. In healthy subjects, EEG activity over the sensorimotor cortex during MI exhibits clear rhythmic characteristics: event-related desynchronization (ERD) in the mu (8--13 Hz) and beta (13--30 Hz) bands is typically observed over the hemisphere contralateral to the imagined limb movement, whereas event-related synchronization (ERS) appears over the ipsilateral hemisphere. This spatiotemporal pattern also shows a relatively stable ROI distribution over the sensorimotor area\cite{Pfurtscheller2001}. For local spatiotemporal modeling, EEGNet\cite{Lawhern2018} and ShallowConvNet\cite{Schirrmeister2017} extract task-relevant features through a compact convolutional architecture and a shallow convolutional pathway, respectively. For temporal and time-frequency modeling, IFNet\cite{Wang2023IFNet} strengthens the coupled representation of different frequency components through interactive frequency convolution. EEGConformer\cite{Song2023EEGConformer} combines convolutional structures with a Transformer to introduce global dependency modeling while preserving local spatiotemporal feature extraction. MSCFormer\cite{Zhao2025MSCFormer} and DBConformer\cite{Wang2025DBConformer} further enhance representation capability from the perspectives of multi-scale modeling and dual-branch parallel spatiotemporal modeling, respectively. Building on these advances, SlimSeiz\cite{Lu2024SlimSeiz} incorporates Mamba-based state-space modeling into a lightweight convolutional backbone. For target-domain adaptation, SSTDA\cite{Chen2025SSTDA} validates the effectiveness of pseudo-label strategies in EEG decoding by jointly exploiting source-domain labels and target-domain pseudo-labels. SSAS\cite{Liu2026SSAS} explicitly selects source domains to preserve more transferable knowledge and alleviate negative transfer. UA-DANN\cite{Shen2025UADAAN} introduces uncertainty awareness into a dynamic adversarial adaptation framework to improve reliability and robustness during domain transfer.

Despite this progress in healthy population, cross-subject MI decoding in stroke patients still faces fundamental challenges because stroke EEG deviates markedly from healthy physiological patterns\cite{11309705, 11429349}. On the one hand, lesions induce rhythm reorganization in the sensorimotor cortex, manifested as abnormal increases in low-frequency slow waves, attenuated or even reversed ERD/ERS amplitudes in the mu/beta rhythms, and disrupted interhemispheric balance, with enhanced compensatory activation in the unaffected hemisphere and dominant inhibitory activity in the affected hemisphere\cite{Kancheva2023Association,Xu2022Time-Varying, Rustamov2022}. On the other hand, local abnormal disturbances, such as pathological spike-and-slow-wave activity and focal slowing, are superimposed on rhythm modulation related to motor intention, so EEG signals contain both neurophysiological responses and pathological noise. In addition, differences in lesion location, lesion size, and impairment severity further amplify the joint-distribution shift across patients, making it difficult for models trained under conventional assumptions to generalize directly\cite{Kaiser2012,Tangwiriyasakul2014,StrokeSMRReview2023}. Existing domain adaptation methods mainly rely on prediction confidence or statistical distribution alignment while overlooking the neurophysiological nature of EEG signals\cite{11328767}. In stroke patients, highly confident predictions may not originate from genuine motor intention but instead from pathological noise, such as low-frequency drift or abnormal rhythms\cite{Keser2022Electroencephalogram}. Such noise can exhibit statistical pseudo-certainty, thereby misleading pseudo-label generation and causing cumulative error propagation.

To address pathological temporal mismatch and pseudo-label unreliability, this paper proposes PA-TCNet for cross-subject MI decoding in stroke patients. The central idea is that stroke EEG should not be treated as a temporally homogeneous process. Instead, explicit decomposition of pathological rhythms, especially the separation of slow-wave background and transient disturbances, should be combined with neurophysiological priors from the sensorimotor ROI, such as class-related ERD/ERS templates. On this basis, PA-TCNet improves adaptation through pathology-aware temporal modeling and physiologically constrained pseudo-label calibration. 

The main contributions are summarized as follows:
\begin{enumerate}
\item PA-TCNet, a pathology-aware adaptation framework for cross-subject MI decoding in stroke patients, is proposed to unify pathological dynamic representation and physiology-guided target calibration within a single training framework.
\item Pathology-aware Rhythmic State Mamba (PRSM) module: an explicit dual-timescale Mamba-based temporal model, is developed to jointly capture slowly varying rhythmic trends and transient pathological details in stroke EEG.
\item Physiology-Guided Target Calibration (PGTC) module: an ROI-based physiological template constraint is introduced to screen and dynamically update target-domain pseudo-labels, thereby improving the reliability of unlabeled target-domain adaptation and reducing cumulative pseudo-labeling errors.

\end{enumerate}

The remainder of this paper is organized as follows. Section~\ref{sec:method} presents the details of the proposed method. Section~\ref{sec:exp} describes the experimental setup, including datasets, preprocessing, evaluation protocol, and implementation details. Section~\ref{sec:results} reports the results and discussion, including overall performance, ablation analysis, and mechanism-oriented interpretation. Section~\ref{sec:conclusion} concludes the paper.

\section{Method}
\label{sec:method}

\begin{figure}[!t]
\centering
\includegraphics[width=\linewidth]{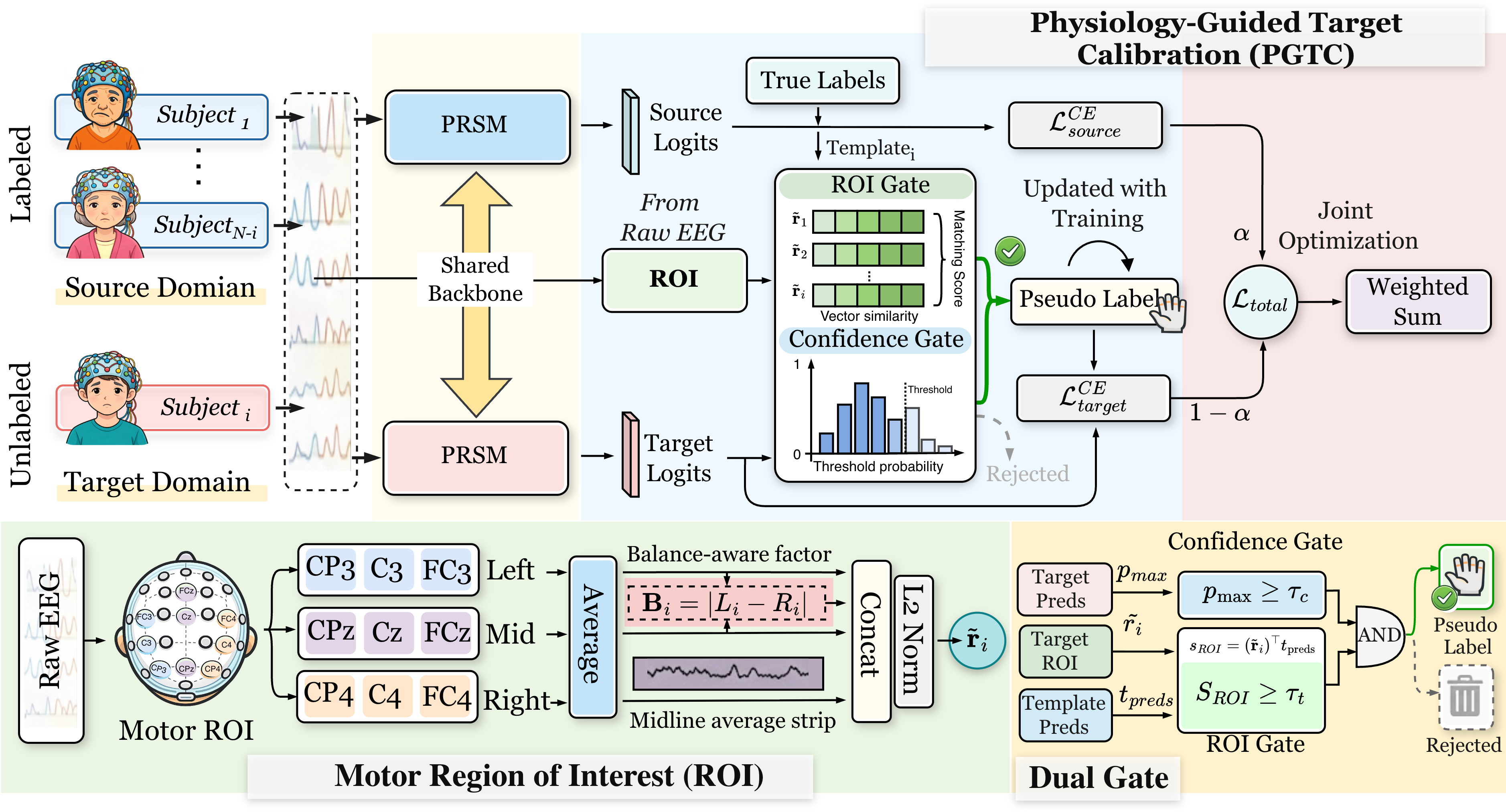}
\caption{Overall framework of PA-TCNet. The network first extracts local sensorimotor spatiotemporal patterns, then performs pathology-aware temporal calibration through PRSM, and finally refines target-domain pseudo-labels with PGTC for joint adaptation.}
\label{fig:framework}
\end{figure}

\begin{figure}[!t]
\centering
\includegraphics[width=\linewidth]{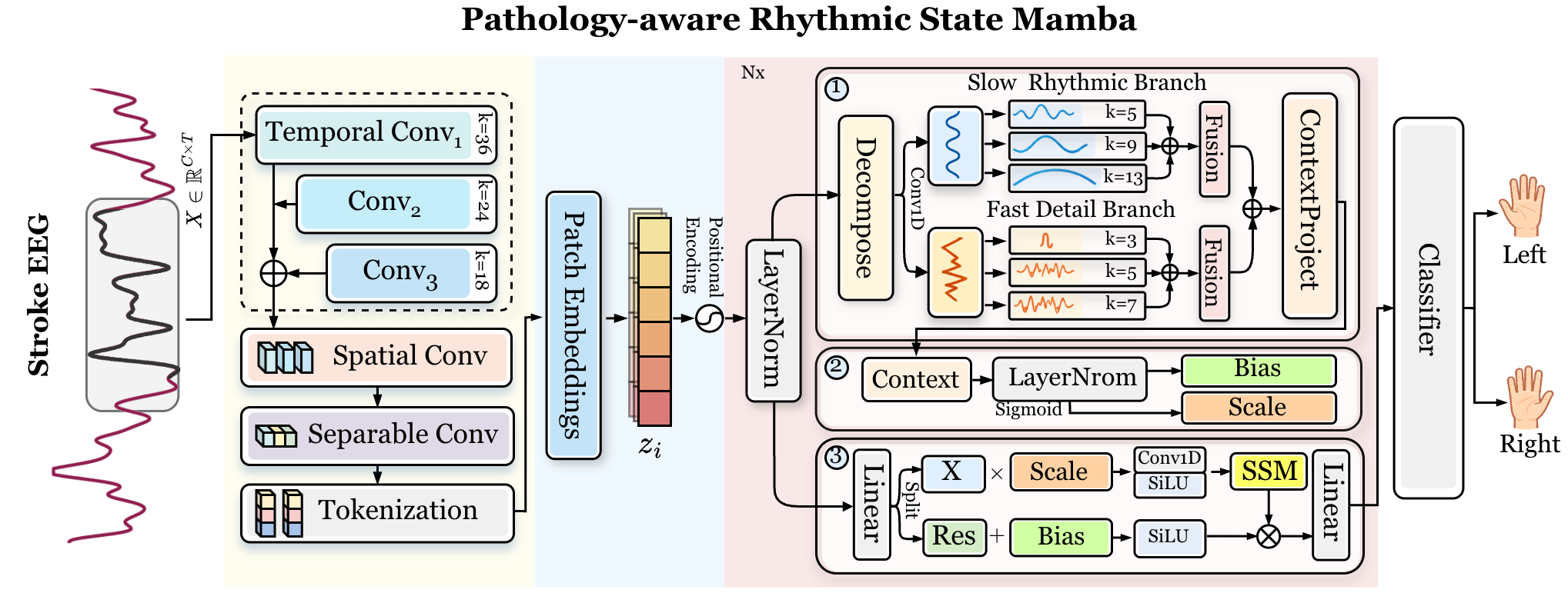}
\caption{Architecture of the Pathology-aware Rhythmic State Mamba (PRSM) module. The module decomposes temporal tokens into slow rhythmic context and fast transient details, fuses both branches into a pathology-aware context, and injects that context into selective state propagation.}
\label{fig:prsm}
\end{figure}

The proposed PA-TCNet consists of two coordinated components: PRSM and PGTC. PRSM aims to model the abnormal temporal dynamics of stroke EEG and produce pathology-aware temporal representations, whereas PGTC calibrates target-domain pseudo-labels by combining prediction confidence with source-domain physiological templates extracted from the sensorimotor ROI. Given labeled source-domain EEG and unlabeled target-domain EEG, the model first extracts local spatiotemporal patterns and organizes them into temporal tokens, then performs temporal modeling and class prediction through PRSM. For target-domain samples, PGTC further screens and dynamically updates pseudo-labels to provide more reliable adaptation supervision. The overall framework and the PRSM block are illustrated in Figs.~\ref{fig:framework} and~\ref{fig:prsm}, respectively.

\subsection{Problem Formulation}
Cross-subject MI decoding in stroke patients can be formulated as an adaptation problem from labeled source-domain data to unlabeled target-domain data. The source and target domains are denoted by:
\begin{equation}
\mathcal{D}_s=\{(X_i^s,Y_i^s)\}_{i=1}^{N_s}, \qquad
\mathcal{D}_t=\{X_j^t\}_{j=1}^{N_t},
\end{equation}
where $X_i^s$ and $Y_i^s$ denote the $i$-th source-domain EEG sample and its label, respectively; $X_j^t$ denotes the $j$--th unlabeled target-domain EEG sample; and $N_s$ and $N_t$ are the numbers of source and target samples.

For an EEG trial $X \in \mathbb{R}^{C\times T}$, $C$ denotes the number of EEG channels and $T$ denotes the number of temporal sampling points in each channel. The $n$-th sample can be represented as $X_n=(x_{n,1},x_{n,2},\dots,x_{n,C})$, where $x_{n,c}\in\mathbb{R}^{T}$ denotes the temporal sequence of the $c$-th channel. The corresponding label $y_n$ is assigned to the entire EEG trial rather than to any individual channel.

In cross-subject stroke MI decoding, the source and target domains typically satisfy:
\begin{equation}
P(X^s,Y^s)\neq P(X^t,Y^t),
\end{equation}
which indicates a significant joint-distribution shift. This shift is reflected not only in the statistical properties of EEG signals across patients, but also in the inconsistency of class-discriminative structures.

The goal of this study is therefore to learn a target-oriented decoding function:
\begin{equation}
h = g_{\phi}\circ f_{\theta},
\end{equation}
where $f_{\theta}$ denotes the feature encoder, $g_{\phi}$ denotes the classifier, and $\circ$ denotes the Hadamard product. The model should simultaneously reduce cross-subject temporal representation mismatch under stroke pathology and improve the reliability of target-domain pseudo-supervision.

\subsection{Pathology-aware Rhythmic State Mamba (PRSM)}

To reduce the representation mismatch caused by abnormal temporal dynamics, PRSM, as shown in Fig.~\ref{fig:prsm}, explicitly distinguishes slowly varying rhythmic background from fast transient disturbances and injects the resulting rhythmic context into selective state propagation. Given an input trial $X\in\mathbb{R}^{C\times T}$, PRSM first generates spatiotemporal token sequences, then performs rhythmic decomposition, multi-scale context fusion, and context-guided state modeling to obtain the pathology-aware representation $Z^{\ast}$.

\subsubsection{Local Sensorimotor Pattern Encoding}

The preprocessed EEG is first mapped into a token sequence by a local sensorimotor pattern encoder. Let $\{\mathcal{T}_m\}_{m=1}^{M}$ denote a set of local temporal filters with different receptive fields. Temporal pattern extraction is expressed as:
\begin{equation}
F_{\mathrm{temp}}=\mathcal{A}_t\!\left(\{\mathcal{T}_m(X)\}_{m=1}^{M}\right),
\end{equation}
where $\mathcal{A}_t(\cdot)$ denotes the aggregation operator over multi-scale temporal responses. A cross-channel spatial mapping $\mathcal{S}(\cdot)$ is then used to model coordinated activation in the sensorimotor area, followed by a compact fusion mapping $\mathcal{F}(\cdot)$ to remove redundant coupling:
\begin{equation}
F_{\mathrm{emb}}=\mathcal{F}\!\big(\mathcal{S}(F_{\mathrm{temp}})\big)
\end{equation}
After temporal compression and serialization, the model obtains a spatiotemporal token sequence:
\begin{equation}
Z=[z_1,z_2,\dots,z_L]\in\mathbb{R}^{L\times D},
\end{equation}
where $L$ is the number of tokens and $D$ is the embedding dimension.

\subsubsection{Rhythmic Context Construction}

To explicitly model pathological dynamics across different timescales in stroke MI-EEG, PRSM performs temporal rhythmic decomposition. Stroke EEG is often accompanied by enhanced low-frequency slow-wave activity. Let the normalized token sequence be $\tilde{Z}$ and let $\mathcal{L}(\cdot)$ denote a feature-level low-frequency extraction operator. The slowly varying rhythmic component $Z_{\mathrm{low}}$ and the local transient component $Z_{\mathrm{high}}$ are defined as:
\begin{equation}
Z_{\mathrm{low}}=\mathcal{L}(\tilde{Z}), \qquad
Z_{\mathrm{high}}=\tilde{Z}-Z_{\mathrm{low}}
\end{equation}
This decomposition enables the model to isolate abnormal low-frequency drift from transient pathological fluctuations. Multi-scale temporal mappings are then applied to the slow-rhythm branch and the fast-detail branch to produce $F_{\mathrm{low}}$ and $F_{\mathrm{high}}$, respectively. The rhythmic context is finally defined as:
\begin{equation}
C_r=\mathcal{H}\!\left(F_{\mathrm{low}},F_{\mathrm{high}}\right)
+\mathcal{A}_r\!\left(F_{\mathrm{low}},F_{\mathrm{high}}\right),
\end{equation}
where $\mathcal{H}(\cdot,\cdot)$ denotes the context fusion mapping and $\mathcal{A}_r(\cdot,\cdot)$ denotes a residual aggregation term.

\subsubsection{Context-guided Selective State Modeling}

The rhythmic context $C_r$ directly participates in state-propagation parameterization. Given the normalized token sequence $\tilde{Z}$, an input projection first generates the main branch and the residual branch:
\begin{equation}
[U,R]=\mathrm{Split}\!\left(W_{\mathrm{in}}\tilde{Z}\right), \qquad
U,R\in\mathbb{R}^{L\times D_i},
\end{equation}
where $D_i$ is the internal state dimension. The rhythmic context then produces an adaptive scaling term and an additive bias term for both branches:
\begin{equation}
S=\sigma(W_s C_r), \qquad B=W_b C_r,
\end{equation}
\begin{equation}
\bar{U}=U\odot(1+S), \qquad \bar{R}=R+B
\end{equation}
Through this operation, pathology-related rhythmic information modulates both the input drive and residual gating in a context-dependent manner.

After local temporal pre-encoding, $\bar{U}$ is fed into the selective state-space model. At the $l$-th position, the state recursion is written as:
\begin{equation}
h_l=\bar{A}_l h_{l-1}+\bar{B}_l u_l, \qquad
o_l=\bar{C}_l h_l + D u_l,
\end{equation}
where $u_l$ is the input at the $l$-th position of $\bar{U}$ and $\bar{A}_l$, $\bar{B}_l$, and $\bar{C}_l$ are input-dependent selective state parameters. The final output representation is:
\begin{equation}
z_l^{\ast}=W_{\mathrm{out}}\!\left(o_l\odot\varphi(\bar{r}_l)\right),
\end{equation}
where $\varphi(\cdot)$ is a nonlinear activation function and $\bar{r}_l$ is the gating term at the $l$-th position of $\bar{R}$. Stacking all positions yields:
\begin{equation}
Z^{\ast}=\mathrm{PRSM}(Z;C_r)
\end{equation}
In this way, slow rhythmic background and fast transient disturbances are fused into $C_r$ and jointly regulate state updates and residual pathways, thereby improving representation stability under abnormal rhythmic drift, local transient perturbations, and cross-subject temporal heterogeneity.

\subsection{Physiology-Guided Target Calibration (PGTC)}
As shown in Fig.~\ref{fig:framework}, PGTC is proposed to enforce a joint criterion of semantic confidence and physiological plausibility. It first derives class-specific ROI templates from the source domain, and then combines prediction probabilities with ROI-template matching to filter and dynamically update pseudo-labels for the target domain.

\subsubsection{Source ROI Template Construction}

PGTC first constructs class-related ROI physiological templates in the source domain. Let $\mathcal{R}(\cdot)$ denote the ROI feature extractor. For any source-domain sample $X_i^s$, its ROI representation is:
\begin{equation}
r_i^s=\mathcal{R}(X_i^s)\in\mathbb{R}^{d_r},
\end{equation}
where $d_r$ is the dimension of the ROI feature vector. In this work, $\mathcal{R}(\cdot)$ is built from fixed sensorimotor electrodes. The left ROI, middle ROI, and right ROI are composed of $\{\mathrm{FC3},\mathrm{C3},\mathrm{CP3}\}$, $\{\mathrm{FCz},\mathrm{Cz},\mathrm{CPz}\}$, and $\{\mathrm{FC4},\mathrm{C4},\mathrm{CP4}\}$, respectively. For the $i$-th source-domain sample, the trial-level average temporal sequences of the three groups are denoted by $L_i^s$, $M_i^s$, and $R_i^s$. A bilateral asymmetry term is then defined as:
\begin{equation}
B_i^s=\left|L_i^s-R_i^s\right|
\end{equation}
The source ROI representation is obtained by concatenating the left, right, asymmetry, and middle terms and then applying $L_2$ normalization:
\begin{equation}
r_i^s=\operatorname{Norm}_{2}\!\left([L_i^s,R_i^s,B_i^s,M_i^s]\right)
\end{equation}

Given the index set of source samples from the $k$-th class, $\mathcal{I}_k=\{i\,|\,Y_i^s=k\}$, the class template is defined as:
\begin{equation}
T_k=\operatorname{Norm}_{2}\!\left(\frac{1}{|\mathcal{I}_k|}\sum_{i\in\mathcal{I}_k}r_i^s\right), \qquad
k\in\{1,\dots,K\},
\end{equation}
which provides the prototype of the $k$-th MI class in ROI space. To characterize intra-class physiological variability, an adaptive template-matching threshold is further defined through cosine similarity:
\begin{equation}
\begin{aligned}
\delta_k=\max\Big(&\delta_{\min}d,\
\frac{1}{|\mathcal{I}_k|}\sum_{i\in\mathcal{I}_k}\mathrm{sim}(r_i^s,T_k) \\
&-\operatorname{Std}\!\big(\{\mathrm{sim}(r_i^s,T_k)\}_{i\in\mathcal{I}_k}\big)\Big),
\end{aligned}
\end{equation}
where $\mathrm{sim}(\cdot,\cdot)$ denotes cosine similarity and $\delta_{\min}$ is the lower bound of the threshold. The resulting source template set is:
\begin{equation}
\mathcal{T}=\{(T_k,\delta_k)\}_{k=1}^{K}
\end{equation}
In subsequent target calibration, $T_k$ provides the class-related physiological prototype and $\delta_k$ defines the acceptable level of physiological consistency.

\subsubsection{Pseudo-label Calibration}

After the source ROI template set $\mathcal{T}$ is constructed, PGTC calibrates the target pseudo-labels. For a target-domain sample $X_j^t$, the classifier first outputs class probabilities and a candidate class, while the ROI representation is extracted simultaneously:
\begin{equation}
\begin{aligned}
p_j&=g_{\phi}(f_{\theta}(X_j^t))\in\mathbb{R}^{K},\\
\hat{y}_j&=\arg\max_{k} p_{j,k}, \\
r_j^t&=\mathcal{R}(X_j^t)
\end{aligned}
\end{equation}
The acceptance condition is then defined by combining semantic confidence and physiological matching:
\begin{equation}
\max_k p_{j,k}>\tau_p, \qquad
\mathrm{sim}(r_j^t,T_{\hat{y}_j})>\delta_{\hat{y}_j},
\end{equation}
where $\tau_p$ is the confidence threshold and $\delta_{\hat{y}_j}$ is the class-specific physiological threshold. The calibrated pseudo-label is defined as:
\begin{equation}
\tilde{y}_j^t=
\begin{cases}
\operatorname{onehot}(\hat{y}_j), & \text{if } X_j^t \text{ satisfies the joint criterion},\\
\mathbf{0}, & \text{otherwise}.
\end{cases}
\end{equation}
Here, $\mathbf{0}$ means that the sample is excluded from target-domain supervision at the current iteration. PGTC further refreshes $\tilde{y}_j^t$ dynamically during training, so the target supervision set becomes:
\begin{equation}
\tilde{\mathcal{D}}_t=\{(X_j^t,\tilde{y}_j^t)\}_{j=1}^{N_t}
\end{equation}
This formulation transforms target-domain pseudo-supervision into a joint constraint process wherein class probabilities propose candidate labels, ROI templates assess their physiological plausibility, and dynamic updates suppress the early mislabel accumulation.

\subsection{Optimization Strategy}

With pathology-aware representations and calibrated target supervision, model training is formulated as a joint optimization problem that combines explicit source-domain supervision with constrained target-domain pseudo-supervision. Because target pseudo-labels are unstable at the beginning of training, a two-stage strategy is adopted: source-only warm-up followed by joint adaptation.

\subsubsection{Warm-up}

During warm-up, only labeled source-domain samples are used to optimize the encoder $f_{\theta}$ and classifier $g_{\phi}$. The objective is:
\begin{equation}
\mathcal{L}_{\mathrm{src}}=
\frac{1}{N_s}\sum_{i=1}^{N_s}
\ell\!\left(g_{\phi}(f_{\theta}(X_i^s)),Y_i^s\right),
\end{equation}
where $\ell(\cdot,\cdot)$ denotes the cross-entropy loss. This stage avoids unstable target predictions from interfering with early parameter updates.

\subsubsection{Joint Adaptation}

After warm-up, the model introduces PGTC-calibrated target pseudo-labels for joint adaptation. The target-domain loss is defined as:
\begin{equation}
\mathcal{L}_{\mathrm{tgt}}=
\frac{1}{N_t}\sum_{j=1}^{N_t}
\mathbf{1}\!\left(\tilde{y}_j^t\neq\mathbf{0}\right)
\ell\!\left(g_{\phi}(f_{\theta}(X_j^t)),\tilde{y}_j^t\right),
\end{equation}
where $\mathbf{1}(\cdot)$ denotes the indicator function. The overall objective is:
\begin{equation}
\mathcal{L}=\alpha\mathcal{L}_{\mathrm{src}}+(1-\alpha)\mathcal{L}_{\mathrm{tgt}},
\end{equation}
where $\alpha\in(0,1)$ controls the trade-off between source-domain supervision and calibrated target-domain pseudo-supervision. After each training round, the model recomputes $p_j$ and $r_j^t$ and refreshes $\tilde{\mathcal{D}}_t$ dynamically so that the target supervision evolves together with representation quality.

\section{Experiments}
\label{sec:exp}

\subsection{Datasets}
\textbf{XW-Stroke:} The XW-Stroke dataset\cite{Liu2024XWStroke} was collected at Xuanwu Hospital, Capital Medical University. The original dataset contains 50 acute ischemic stroke patients (39 males and 11 females). All subjects performed left-hand and right-hand MI tasks, and EEG was recorded with 30 channels under the international 10--20 system at 500 Hz. To reduce clinical heterogeneity that was not directly aligned with the cross-subject modeling objective, a well-defined patient subcohort was further selected according to predefined clinical inclusion criteria: right-handedness, first-ever stroke, NIHSS score below 10, and disease duration below 10 years. These criteria were determined solely by clinical attributes and were independent of the model training and classification outcomes. The final cohort included 24 patients, indexed as 2, 5, 8, 9, 11, 12, 14, 17, 21, 23, 24, 26, 27, 28, 30, 32, 33, 37, 38, 43, 44, 47, 49, and 50.

\textbf{2019-Stroke:} The 2019-Stroke dataset was collected by Jia \textit{et al.}\cite{Jia2019}. It contains 15 stroke patients, each performing left-hand and right-hand MI tasks, with both paretic-side and non-paretic-side imagination included in the original acquisition setting. EEG was recorded with 63 channels under the international 10--10 system at 512 Hz. All publicly available subjects were used without additional subject-level exclusion.

\begin{table}[!t]
\caption{Summary of dataset protocols and model input specifications used in this study.}
\label{tab:datasets}
\centering
\small
\resizebox{\linewidth}{!}{%
\begin{tabular}{l l l c c c}
\toprule
\textbf{Dataset} & \textbf{Task} & \textbf{System} & \textbf{Sampling rate} & \textbf{Trials per subject} & \textbf{Model input shape} \\
\midrule
XW-Stroke & Left- vs. right-hand MI & 10--20 & 500 Hz & 40 (20 per class) & $30\times1000$ \\
2019-Stroke & Left- vs. right-hand MI & 10--10 & 512 Hz & 80 (40 per class) & $63\times1708$ \\
\bottomrule
\end{tabular}
}
\end{table}

\subsection{Preprocessing}

To preserve MI-related rhythms while suppressing interference, the raw EEG signals were first band-pass filtered between 8 and 30 Hz and then downsampled to 250 Hz. All channels were subsequently re-referenced by common average referencing and baseline corrected to reduce common noise and slow drift. Finally, independent component analysis (ICA) was applied to separate and remove artifact-related components.

\subsection{Evaluation Protocol and Metrics}
In this study, Leave-one-subject-out (LOSO) protocol was adopted to evaluate model performance. In each round, one stroke patient was selected as the target subject and all remaining subjects were treated as the source domain. The model was trained on the source-domain data and evaluated on the held-out target subject. This procedure was repeated until every subject had been used once as the target domain. To avoid data leakage, subject-level splitting was performed before training so that the target subject did not participate in source training with labels and no labeled information was shared across source and target domains.

The final performance was obtained by averaging all evaluation metrics across all LOSO rounds. Specifically, classification accuracy (Accuracy), Cohen’s kappa coefficient, recall, precision, and F1-score were used to assess cross-subject generalization performance. In addition, Wilcoxon signed-rank tests were used to compute $p$-values for significance analysis between competing methods.

\subsection{Implementation Details and Hyperparameter Settings}

The proposed PA-TCNet was implemented via PyTorch under Python 3.12. Training used the Adam optimizer with an initial learning rate of 0.001, a weight decay of 0.001, and a maximum of 200 epochs. All experiments were conducted on an AMD EPYC CPU and an NVIDIA RTX 3090 GPU. The architectural hyperparameters of PA-TCNet are summarized in Table~\ref{tab:model-hparams}, and the dataset-specific default adaptation settings are listed in Table~\ref{tab:adapt-hparams}.

\begin{table}[!t]
\caption{Detailed architectural hyperparameters of PA-TCNet.}
\label{tab:model-hparams}
\centering
\small
\begin{tabular}{ll}
\toprule
\textbf{Parameter} & \textbf{Value} \\
\midrule
Embedding size & 30 \\
Encoder depth & 2 \\
Temporal filters per branch & 10 \\
Spatial multiplier & 3 \\
Temporal kernel sizes & $\{36,24,18\}$ \\
Pooling size & 8 \\
Frequency kernel setting & $\{5,9,13\},\{3,5,7\}$ \\
\bottomrule
\end{tabular}
\end{table}

\begin{table}[!t]
\caption{Dataset-specific default adaptation hyperparameters of PA-TCNet.}
\label{tab:adapt-hparams}
\centering
\small
\begin{tabular}{lcc}
\toprule
\textbf{Hyperparameter} & \textbf{XW-Stroke} & \textbf{2019-Stroke} \\
\midrule
Source loss weight $\alpha$ & 0.98 & 0.95 \\
Confidence threshold $\tau_p$ & 0.60 & 0.60 \\
ROI threshold floor $\delta_{\min}$ & 0.50 & 0.45 \\
Warm-up epochs $E_w$ & 25 & 10 \\
\bottomrule
\end{tabular}
\end{table}

\section{Results and Discussion}
\label{sec:results}

\subsection{Overall Performance}

To evaluate the effectiveness of proposed PA-TCNet on cross-subject MI classification in stroke patients, this study compared it with representative baselines on XW-Stroke and 2019-Stroke. The baselines included three CNN-based models, EEGNet\cite{Lawhern2018}, ShallowConvNet\cite{Schirrmeister2017}, and IFNet\cite{Wang2023IFNet}; three Transformer-based models, EEGConformer\cite{Song2023EEGConformer}, MSCFormer\cite{Zhao2025MSCFormer}, and DBConformer\cite{Wang2025DBConformer}; one Mamba-based model, SlimSeiz\cite{Lu2024SlimSeiz}; and three transfer or adaptation methods, UA-DANN\cite{Shen2025UADAAN}, SSTDA\cite{Chen2025SSTDA}, and SSAS\cite{Liu2026SSAS}.

As reported in Table~\ref{tab:overall_xw} and Table~\ref{tab:overall_2019}, PA-TCNet achieved the best performance on both datasets. On XW-Stroke, it improved the best competing method by 3.96\%. On 2019-Stroke, the gain reached 7.67\%. Relative to classical convolutional models, PA-TCNet improved accuracy by 5.00\% and 4.27\% over EEGNet and ShallowConvNet on XW-Stroke, and by 16.92\% and 15.50\% on 2019-Stroke. The consistent advantage over Transformer-based and adaptation-based baselines indicates that jointly modeling pathological temporal dynamics and physiology-guided pseudo-supervision yields stronger cross-subject generalization under stroke-specific distribution shifts.

Overall, PA-TCNet obtained the highest accuracy and kappa on both datasets. The significance analysis further showed that, except for the comparisons with UA-DANN, SSTDA, and SSAS on 2019-Stroke, the performance differences between PA-TCNet and the remaining baselines reached statistical significance ($p<0.05$). These results confirm that PA-TCNet maintains a stable performance advantage across different stroke data distributions.

\begin{table}[!t]
\caption{Comparison of MI classification performance on the XW-Stroke dataset.}
\label{tab:overall_xw}
\centering
\scriptsize
\renewcommand{\arraystretch}{1.1}
\resizebox{\linewidth}{!}{%
\begin{tabular}{lccccc}
\toprule
\textbf{Method} & \textbf{Accuracy (\%)} & \textbf{Recall (\%)} & \textbf{Precision (\%)} & \textbf{F1-Score (\%)} & \textbf{Kappa} \\
\midrule
EEGNet\cite{Lawhern2018}$^{**}$                 & 61.56 ± 06.53 & 61.87 ± 13.83 & 62.00 ± 07.90 & 61.11 ± 08.37 & 0.231 ± 0.131 \\
ShallowConvNet\cite{Schirrmeister2017}$^{**}$   & 62.29 ± 05.44 & 65.42 ± 14.14 & 62.32 ± 05.92 & 62.68 ± 08.57 & 0.246 ± 0.109 \\
IFNet\cite{Wang2023IFNet}$^{*}$                 & 61.88 ± 07.15 & 54.38 ± 17.22 & 67.20 ± 11.94 & 57.23 ± 13.25 & 0.237 ± 0.143 \\
EEGConformer\cite{Song2023EEGConformer}$^{**}$  & 61.67 ± 04.66 & 65.00 ± 17.38 & 62.83 ± 06.48 & 61.68 ± 09.43 & 0.233 ± 0.093 \\
MSCFormer\cite{Zhao2025MSCFormer}$^{**}$        & 58.96 ± 05.30 & 61.88 ± 26.88 & 61.87 ± 10.33 & 56.89 ± 14.89 & 0.179 ± 0.106 \\
DBConformer\cite{Wang2025DBConformer}$^{**}$    & 61.15 ± 04.89 & 62.29 ± 26.34 & 67.49 ± 14.54 & 58.45 ± 14.19 & 0.223 ± 0.098 \\
SlimSeiz\cite{Lu2024SlimSeiz}$^{**}$            & 58.75 ± 05.64 & 53.12 ± 22.68 & 63.41 ± 11.21 & 53.70 ± 13.62 & 0.175 ± 0.113 \\
UA-DANN\cite{Shen2025UADAAN}$^{**}$             & 58.33 ± 03.86 & 49.17 ± 15.72 & 60.68 ± 04.52 & 52.81 ± 09.59 & 0.167 ± 0.077 \\
SSTDA\cite{Chen2025SSTDA}$^{**}$                & 62.60 ± 05.57 & 54.79 ± 18.00 & 67.45 ± 11.04 & 57.98 ± 10.33 & 0.252 ± 0.111 \\
SSAS\cite{Liu2026SSAS}$^{**}$                   & 60.42 ± 06.15 & 57.08 ± 18.59 & 59.59 ± 14.19 & 56.98 ± 15.06 & 0.208 ± 0.123 \\
\textbf{PA-TCNet} & \textbf{66.56 ± 06.61} & \textbf{63.12 ± 15.86} & \textbf{69.57 ± 09.55} & \textbf{64.30 ± 10.92} & \textbf{0.331 ± 0.132} \\
\bottomrule
\end{tabular}
}

\vspace{2pt}
\raggedright
\footnotesize{$^{*}$ and $^{**}$ indicate statistically significant differences in Accuracy compared with PA-TCNet at $p<0.05$ and $p<0.01$, respectively.}
\end{table}

\begin{table}[!t]
\caption{Comparison of MI classification performance on the 2019-Stroke dataset.}
\label{tab:overall_2019}
\centering
\scriptsize
\renewcommand{\arraystretch}{1.1}
\resizebox{\linewidth}{!}{%
\begin{tabular}{lccccc}
\toprule
\textbf{Method} & \textbf{Accuracy (\%)} & \textbf{Recall (\%)} & \textbf{Precision (\%)} & \textbf{F1-Score (\%)} & \textbf{Kappa} \\
\midrule
EEGNet\cite{Lawhern2018}$^{*}$                  & 55.83 ± 06.61 & 50.67 ± 31.94 & 61.09 ± 23.42 & 46.35 ± 24.93 & 0.117 ± 0.132 \\
ShallowConvNet\cite{Schirrmeister2017}$^{**}$   & 57.25 ± 07.43 & 54.67 ± 37.06 & 58.06 ± 27.49 & 47.84 ± 26.24 & 0.145 ± 0.149 \\
IFNet\cite{Wang2023IFNet}$^{**}$                & 55.42 ± 04.44 & 52.00 ± 22.01 & 60.62 ± 12.74 & 51.21 ± 13.35 & 0.108 ± 0.089 \\
EEGConformer\cite{Song2023EEGConformer}$^{*}$   & 58.17 ± 04.81 & 58.17 ± 26.39 & 63.73 ± 12.67 & 54.67 ± 16.02 & 0.163 ± 0.096 \\
MSCFormer\cite{Zhao2025MSCFormer}$^{*}$         & 60.08 ± 08.07 & 57.00 ± 35.45 & 66.58 ± 24.90 & 52.05 ± 23.17 & 0.202 ± 0.161 \\
DBConformer\cite{Wang2025DBConformer}$^{**}$    & 56.17 ± 07.90 & 43.00 ± 39.45 & 63.68 ± 30.54 & 38.63 ± 28.89 & 0.123 ± 0.158 \\
SlimSeiz\cite{Lu2024SlimSeiz}$^{*}$             & 57.33 ± 07.50 & 47.50 ± 32.44 & 66.70 ± 15.35 & 46.09 ± 23.24 & 0.147 ± 0.150 \\
UA-DANN\cite{Shen2025UADAAN}                    & 61.75 ± 05.45 & 68.33 ± 18.23 & 62.11 ± 06.61 & 62.86 ± 09.94 & 0.235 ± 0.109 \\
SSTDA\cite{Chen2025SSTDA}                       & 63.50 ± 15.99 & 53.83 ± 35.61 & 58.78 ± 26.91 & 52.20 ± 31.10 & 0.270 ± 0.320 \\
SSAS\cite{Liu2026SSAS}                          & 65.08 ± 20.12 & 68.33 ± 26.72 & 64.27 ± 20.91 & 64.59 ± 23.55 & 0.302 ± 0.402 \\
\textbf{PA-TCNet}    & \textbf{72.75 ± 17.36} & \textbf{71.67 ± 23.62} & \textbf{75.25 ± 18.23} & \textbf{71.50 ± 19.24} & \textbf{0.455 ± 0.347} \\
\bottomrule
\end{tabular}
}

\vspace{2pt}
\raggedright
\footnotesize{$^{*}$ and $^{**}$ indicate statistically significant differences in Accuracy compared with PA-TCNet at $p<0.05$ and $p<0.01$, respectively.}
\end{table}

\subsection{Visualization of Feature Alignment}

\begin{figure}[!t]
\centering
\includegraphics[width=\columnwidth]{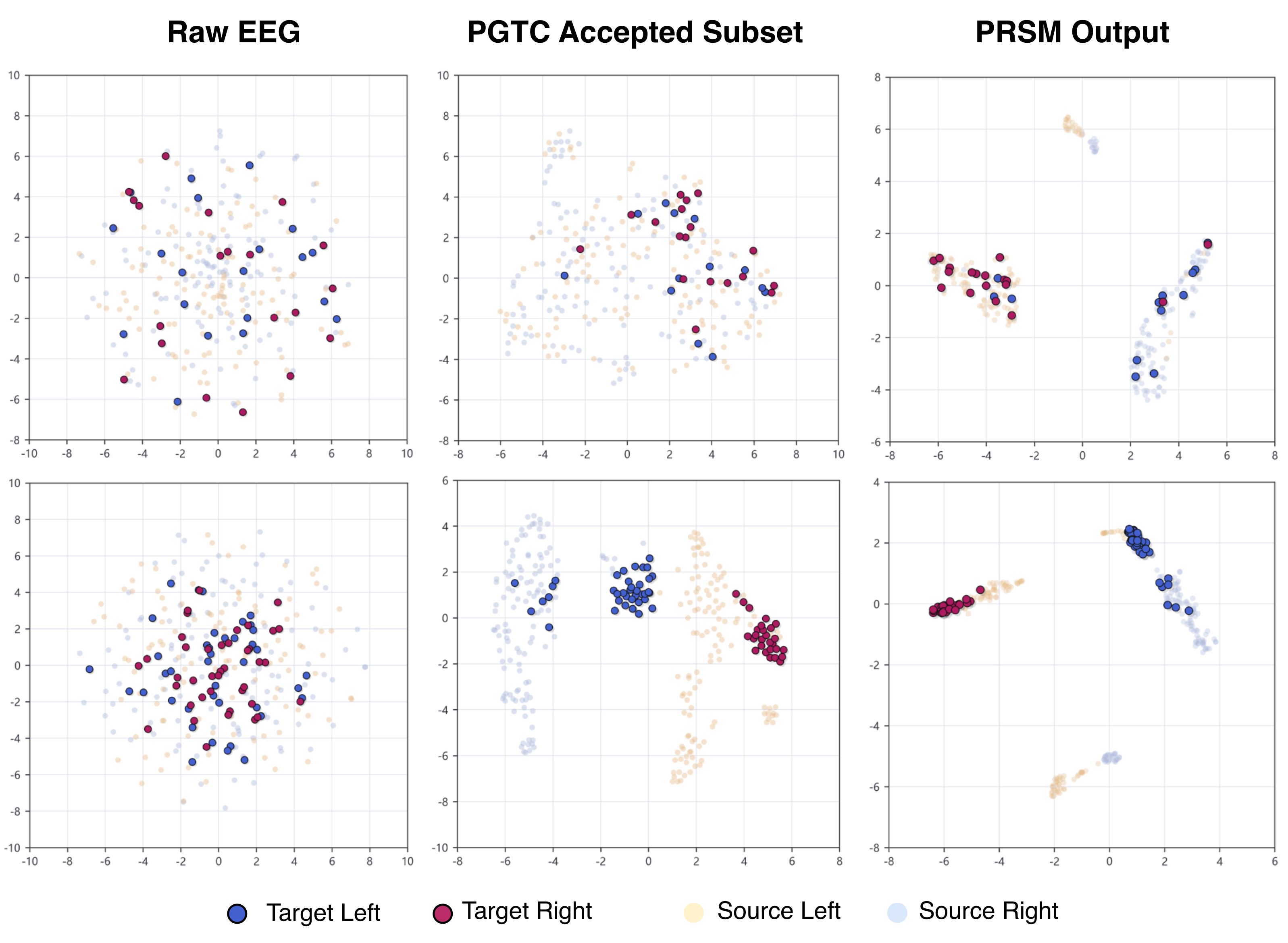}
\caption{t-SNE visualizations of feature distributions on the XW-Stroke (top) and 2019-Stroke (bottom) datasets at different stages of PA-TCNet.}
\label{fig:tsne}
\end{figure}

Fig.~\ref{fig:tsne} provides an intuitive view of the working mechanism of PA-TCNet using t-SNE\cite{Maaten2008}. In the original EEG space, source-domain data (semi-transparent points) and target-domain data (solid points) are irregularly interwoven, indicating severe cross-domain distribution shift and weak class-wise aggregation across patients. After PGTC screening, the accepted target samples begin to contract toward class-related regions with clearer physiological meaning, suggesting that physiology-guided pseudo-labeling effectively filters unreliable target samples while retaining a more discriminative target subset. In the final PRSM output space, the source and target features become better aligned and form more clearly separated left-hand and right-hand MI clusters. This progressive evolution shows that physiology-guided adaptation can map disordered pathological EEG features into a more separable semantic space.

\subsection{Ablation Study}

To verify the contribution of each component, module-level ablation studies are conducted on both datasets, with the results summarizing in Table~\ref{tab:ablation}. Compared with the full model, the Source Only setting decreased by 8.33\% on 2019-Stroke and 2.71\% on XW-Stroke, showing that source supervision alone is insufficient to overcome cross-subject distribution shift in stroke EEG. Removing PGTC caused accuracy drops of 6.08\% and 1.98\%, respectively, indicating that physiology-guided pseudo-label calibration is important for stabilizing target-domain supervision. Removing ROI consistency reduced performance by 5.25\% on 2019-Stroke and 1.66\% on XW-Stroke, while removing dynamic pseudo-label updates caused drops of 5.92\% and 1.25\%, respectively. These observations confirm that physiological template matching and dynamic pseudo-label refresh jointly suppress cumulative pseudo-supervision errors.

Removing rhythmic state calibration reduced the model to a standard Mamba-style temporal state model and led to decreases of 4.83\% on 2019-Stroke and 0.41\% on XW-Stroke. This result supports our hypothesis that stroke EEG exhibits pronounced frequency-scale heterogeneity, namely slowly varying pathological trends together with transient ERD-related perturbations, and that explicitly injecting slowly evolving pathological context into state propagation helps preserve informative motor-intention cues under strong low-frequency drift.

\begin{table}[!t]
\caption{Ablation results of PA-TCNet across different datasets.}
\label{tab:ablation}
\centering
\scriptsize
\resizebox{0.6\columnwidth}{!}{%
\begin{tabular}{llll}
\toprule
\textbf{Dataset} & \textbf{Variant} & \textbf{Accuracy (\%)} & \textbf{Change (\%)} \\
\midrule
\multirow{6}{*}{2019-Stroke}
& \textbf{PA-TCNet} & \textbf{72.75 ± 17.36} & -- \\
& Source Only & 64.42 ± 11.84 & -8.33 \\
& w/o PGTC & 66.67 ± 12.33 & -6.08 \\
& w/o ROI Consistency & 67.50 ± 15.82 & -5.25 \\
& w/o Dynamic Pseudo Update & 66.83 ± 15.66 & -5.92 \\
& w/o Rhythmic State Calibration & 67.92 ± 15.54 & -4.83 \\
\midrule
\multirow{6}{*}{XW-Stroke}
& \textbf{PA-TCNet} & \textbf{66.56 ± 06.61} & -- \\
& Source Only & 63.85 ± 05.86 & -2.71 \\
& w/o PGTC & 64.58 ± 06.80 & -1.98 \\
& w/o ROI Consistency & 64.90 ± 04.36  & -1.66 \\
& w/o Dynamic Pseudo Update & 65.31 ± 08.67 & -1.25 \\
& w/o Rhythmic State Calibration & 66.15 ± 05.64 & -0.41 \\
\bottomrule
\end{tabular}
}
\end{table}

\subsection{Hyperparameter Sensitivity Analysis}

To evaluate the stability of PA-TCNet with respect to key parameters, we analyzed the pseudo-label confidence threshold $\tau_p$, the warm-up duration $E_w$, and the lower bound of the physiological threshold $\delta_{\min}$ on both datasets. We also compared different network depths and embedding dimensions.

\begin{figure}[!t]
\centering
\includegraphics[width=\columnwidth]{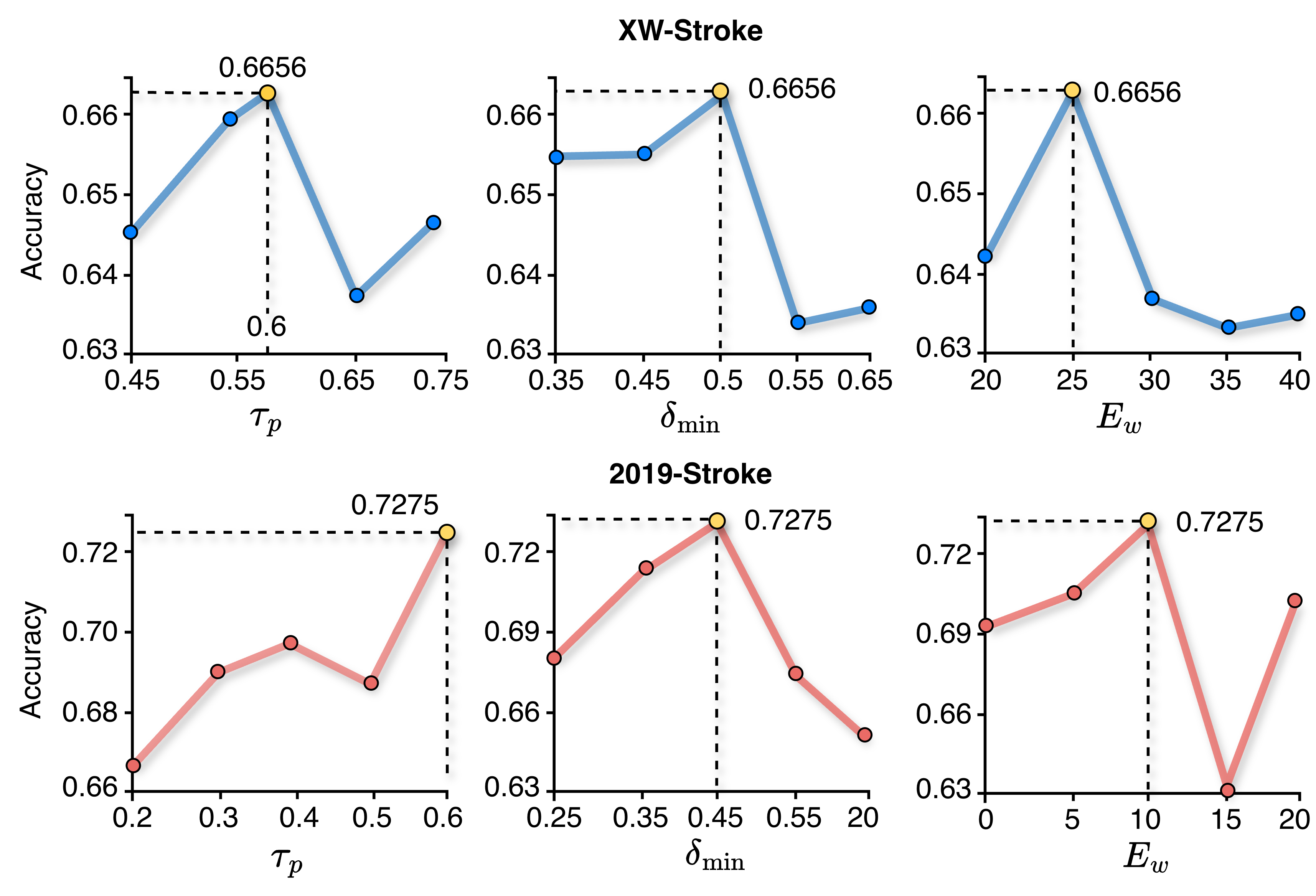}
\caption{Hyperparameter sensitivity analysis of PA-TCNet on XW-Stroke and 2019-Stroke.}
\label{fig:hyper}
\end{figure}

Fig.~\ref{fig:hyper} shows that both $\tau_p$ and $\delta_{\min}$ exhibit an approximately inverted-U performance trend. A smaller $\tau_p$ relaxes pseudo-label activation too much, whereas a larger $\tau_p$ reduces target-domain supervision coverage. This trade-off motivates the default value of 0.60. For $\delta_{\min}$, overly small values make physiological filtering too permissive, while overly large values reject too many genuine stroke samples whose ERD patterns are naturally weakened by pathology. The relatively stable performance region around 0.4--0.6 indicates that the model must balance physiological priors against pathological variability. For warm-up duration, a too-small $E_w$ introduces unstable pseudo-labels too early, whereas a too-large $E_w$ delays target adaptation. The best settings were 25 epochs for XW-Stroke and 10 epochs for 2019-Stroke. In addition, the best-performing model depth was 2 and the best embedding dimension was 30, suggesting that the effective information flow in post-stroke EEG remains relatively low-dimensional and that overly complex models are more prone to overfitting patient-specific noise.

\subsection{Noise Robustness Test}

\begin{figure}[!t]
\centering
\includegraphics[width=\columnwidth]{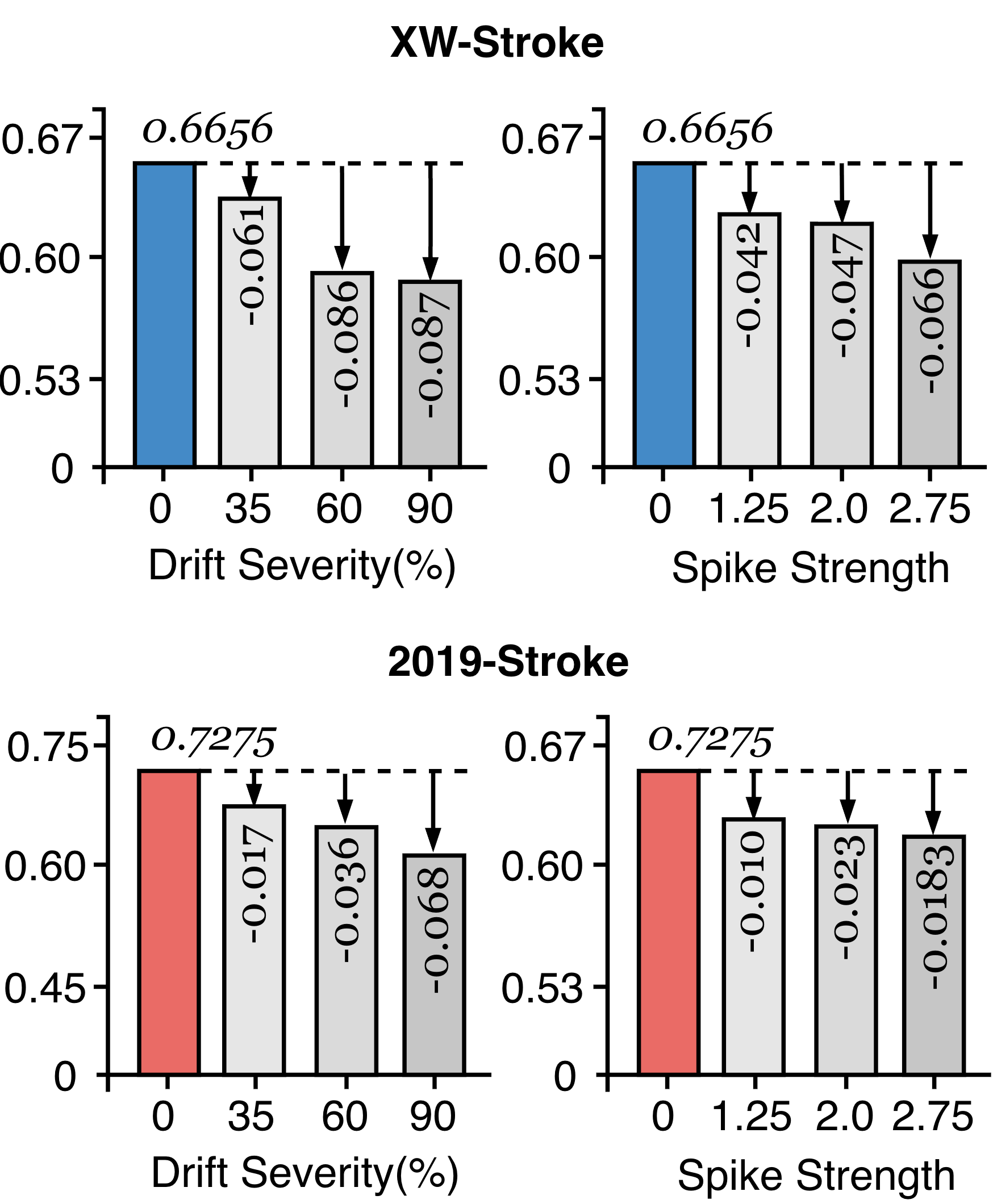}
\caption{Noise robustness of PA-TCNet under drift and spike perturbations.}
\label{fig:noise}
\end{figure}

In real rehabilitation scenarios, stroke patients are often accompanied by limb spasticity, sweating-induced impedance changes, and other sources of signal contamination. To assess robustness, we injected baseline drift that simulates polarization-voltage offset and transient spike noise that simulates muscle spasm into the test data. As shown in Fig.~\ref{fig:noise}, all performance metrics inevitably decreased as disturbance intensity increased, but PA-TCNet exhibited a relatively smooth degradation trend. In particular, the model remained highly resistant to transient spike noise. This behavior can be attributed to the dual-branch decomposition in PRSM, where high-frequency spikes are largely isolated in the fast-detail pathway instead of causing global hidden-state collapse. By contrast, drift noise caused slightly stronger degradation because low-frequency drift is more easily entangled with the intrinsic pathological slow waves of stroke EEG, increasing the difficulty of rhythmic decomposition. Overall, the results indicate that PA-TCNet preserves continuous and interpretable performance evolution under disturbed conditions.

\subsection{Physiological Interpretability Analysis}

Because 2019-Stroke does not provide demographic metadata, this analysis is mainly based on XW-Stroke. Higher decoding performance of PA-TCNet was primarily observed in patients with shorter disease duration and milder neurological deficit: the group with disease duration $\leq 3$ years achieved 68.67\%, compared with 63.06\% for the group with disease duration $>3$ years; the group with NIHSS $\leq 3$ reached 67.50\%, compared with 64.29\% for NIHSS $>3$. Moreover, the five best-performing patients all fell within the range of 1--3 years of disease duration and NIHSS scores between 1 and 3. These results suggest that the model is particularly effective when residual sensorimotor discriminative structure is still preserved in stroke EEG.

This tendency is consistent with the roles of the two proposed modules. PRSM is designed to recover weak lateralized ERD cues when post-stroke $\mu/\beta$ desynchronization is weakened and partially masked by slowly varying pathological rhythms. PGTC incorporates left, midline, right, and asymmetry patterns into the ROI constraint, thereby suppressing high-confidence pseudo-labels that do not match plausible sensorimotor topology. As a result, the gain of PA-TCNet mainly comes from extracting and amplifying residual sensorimotor discriminative structure rather than from purely statistical alignment.

\subsection{Limitations and Future Directions}

Although PA-TCNet was validated on two public stroke EEG datasets, the present study should still be regarded as an initial exploration of stroke-physiology-constrained cross-subject decoding. First, publicly available stroke EEG data remain limited in scale. In particular, XW-Stroke consists of acute-stroke recordings with substantial heterogeneity, and the current study could only validate the method on a screened subcohort rather than on the entire population. The relatively large standard deviations observed in some cross-subject results also suggest that decoding performance remains sensitive to inter-patient variability, which is consistent with the marked heterogeneity of stroke populations in terms of lesion condition, functional impairment, and neurophysiological reorganization. Therefore, the current conclusions still have limited external validity across broader ranges of disease duration, severity, and functional status. Second, because the public datasets do not provide sufficiently complete metadata, this study is not able to perform a more systematic stratified analysis involving lesion hemisphere, lesion location, lesion volume, and finer-grained functional scores, nor could we build a more direct lesion-aware individualized modeling framework. Third, although the method was evaluated on two heterogeneous public datasets, rigorous cross-center validation, stability analysis under different acquisition devices and protocols, and long-term online adaptation in closed-loop rehabilitation settings all remain to be investigated. Finally, the present work uses a fixed sensorimotor ROI template as the physiological prior. This design provides an interpretable and stable weak constraint for target-domain supervision, but it cannot fully capture the complex individualized functional reorganization that occurs after stroke. The practical value of PA-TCNet lies in providing a more stable cross-subject initialization for patient-specific rehabilitation BCIs, rather than replacing clinical assessment or neurological judgment. Future work should therefore combine larger multi-center datasets, richer lesion-level information, online rehabilitation paradigms, more comprehensive evaluation protocols, and lower-burden wearable EEG acquisition settings to further advance clinically usable and deployable stroke MI-BCI systems \cite{he2026,wang2025,11345330}.

\section{Conclusion}
\label{sec:conclusion}

This paper proposed PA-TCNet for cross-subject MI-BCI decoding in stroke patients to address two coupled challenges: pathology-induced temporal representation mismatch and unreliable target-domain pseudo-supervision. PRSM enhances the representation of abnormal temporal dynamics through rhythmic decomposition, context construction, and state modulation, whereas PGTC improves the reliability of target supervision by screening and dynamically updating pseudo-labels using source-domain ROI physiological templates. Experimental results showed that PA-TCNet achieved the best performance under the LOSO protocol and maintained consistent advantages in ablation analysis, hyperparameter sensitivity evaluation, noise robustness testing, and feature visualization. Overall, the study shows that pathology-aware temporal modeling and physiology-guided target refinement synergistically enhance the discriminability, stability, and practical robustness of cross-subject stroke MI decoding. This framework further offers a clinically meaningful approach for integrating neurophysiological priors into deep learning-based BCI systems for post-stroke rehabilitation.

\bibliographystyle{unsrt}
\bibliography{references}

@article{Feigin2022,
 author = {Valery L Feigin and Michael Brainin and Bo Norrving and Sheila Martins and Ralph L Sacco and Werner Hacke and Marc Fisher and Jeyaraj Pandian and Patrice Lindsay},
 title ={World Stroke Organization (WSO): Global Stroke Fact Sheet 2022},
 
 journal = {International Journal of Stroke},
 volume = {17},
 number = {1},
 pages = {18-29},
 year = {2022},
 doi = {10.1177/17474930211065917},
     note ={PMID: 34986727},
 URL = { 
     
         https://doi.org/10.1177/17474930211065917
 
 },
 eprint = { 
     
         https://doi.org/10.1177/17474930211065917
 }
}

@article{Wilson2016,
  author = {Richard D. Wilson and Stephen J. Page and Michael Delahanty and Jayme S. Knutson and Douglas D. Gunzler and Lynne R. Sheffler and John Chae},
  title ={Upper-Limb Recovery After Stroke: A Randomized Controlled Trial Comparing EMG-Triggered, Cyclic, and Sensory Electrical Stimulation},
  
  journal = {Neurorehabilitation and Neural Repair},
  volume = {30},
  number = {10},
  pages = {978-987},
  year = {2016},
  doi = {10.1177/1545968316650278},
      note ={PMID: 27225977},
  URL = { 
      
          https://doi.org/10.1177/1545968316650278  
  },
  eprint = { 
      
          https://doi.org/10.1177/1545968316650278
  }
  }

@article{JIANG2026109680,
  title = {A comprehensive review of deep learning for motor imagery EEG: From healthy subjects to patients},
  journal = {Biomedical Signal Processing and Control},
  volume = {117},
  pages = {109680},
  year = {2026},
  issn = {1746-8094},
  doi = {https://doi.org/10.1016/j.bspc.2026.109680},
  url = {https://www.sciencedirect.com/science/article/pii/S174680942600234X},
  author = {Bin Jiang and Xiangkai Wang and Dongyi He and Siyu Cheng and Maoyu Liao and Duoqian Miao and Qingling Xia and Yun Zhao and Gen Li}
  }

@article{Abiri2019,
  doi = {10.1088/1741-2552/aaf12e},
  url = {https://doi.org/10.1088/1741-2552/aaf12e},
  year = {2019},
  month = {jan},
  publisher = {IOP Publishing},
  volume = {16},
  number = {1},
  pages = {011001},
  author = {Abiri, Reza and Borhani, Soheil and Sellers, Eric W and Jiang, Yang and Zhao, Xiaopeng},
  title = {A comprehensive review of EEG-based brain–computer interface paradigms},
  journal = {Journal of Neural Engineering}
}

@ARTICLE{Pfurtscheller2001,
    author={Pfurtscheller, G. and Neuper, C.},
    journal={Proceedings of the IEEE}, 
    title={Motor imagery and direct brain-computer communication}, 
    year={2001},
    volume={89},
    number={7},
    pages={1123-1134},
    doi={10.1109/5.939829}
}

@article{Lawhern2018,
doi = {10.1088/1741-2552/aace8c},
url = {https://doi.org/10.1088/1741-2552/aace8c},
year = {2018},
month = {jul},
publisher = {IOP Publishing},
volume = {15},
number = {5},
pages = {056013},
author = {Lawhern, Vernon J and Solon, Amelia J and Waytowich, Nicholas R and Gordon, Stephen M and Hung, Chou P and Lance, Brent J},
title = {EEGNet: a compact convolutional neural network for EEG-based brain–computer interfaces},
journal = {Journal of Neural Engineering}
}

@article{Schirrmeister2017,
author = {Schirrmeister, Robin Tibor and Springenberg, Jost Tobias and Fiederer, Lukas Dominique Josef and Glasstetter, Martin and Eggensperger, Katharina and Tangermann, Michael and Hutter, Frank and Burgard, Wolfram and Ball, Tonio},
title = {Deep learning with convolutional neural networks for EEG decoding and visualization},
journal = {Human Brain Mapping},
volume = {38},
number = {11},
pages = {5391-5420},
doi = {https://doi.org/10.1002/hbm.23730},
url = {https://onlinelibrary.wiley.com/doi/abs/10.1002/hbm.23730},
eprint = {https://onlinelibrary.wiley.com/doi/pdf/10.1002/hbm.23730},
year = {2017}
}

@ARTICLE{Wang2023IFNet,
  author={Wang, Jiaheng and Yao, Lin and Wang, Yueming},
  journal={IEEE Transactions on Neural Systems and Rehabilitation Engineering}, 
  title={IFNet: An Interactive Frequency Convolutional Neural Network for Enhancing Motor Imagery Decoding From EEG}, 
  year={2023},
  volume={31},
  number={},
  pages={1900-1911},
  doi={10.1109/TNSRE.2023.3257319}}

@ARTICLE{Song2023EEGConformer,
  author={Song, Yonghao and Zheng, Qingqing and Liu, Bingchuan and Gao, Xiaorong},
  journal={IEEE Transactions on Neural Systems and Rehabilitation Engineering}, 
  title={EEG Conformer: Convolutional Transformer for EEG Decoding and Visualization}, 
  year={2023},
  volume={31},
  number={},
  pages={710-719},
  doi={10.1109/TNSRE.2022.3230250}}

@Article{Zhao2025MSCFormer,
author={Zhao, Wei
and Zhang, Baocan
and Zhou, Haifeng
and Wei, Dezhi
and Huang, Chenxi
and Lan, Quan},
title={Multi-scale convolutional transformer network for motor imagery brain-computer interface},
journal={Scientific Reports},
year={2025},
month={Apr},
day={15},
volume={15},
number={1},
pages={12935},
issn={2045-2322},
doi={10.1038/s41598-025-96611-5},
url={https://doi.org/10.1038/s41598-025-96611-5}
}

@ARTICLE{Wang2025DBConformer,
  author={Wang, Ziwei and Wang, Hongbin and Jia, Tianwang and He, Xingyi and Li, Siyang and Wu, Dongrui},
  journal={IEEE Journal of Biomedical and Health Informatics}, 
  title={DBConformer: Dual-Branch Convolutional Transformer for EEG Decoding}, 
  year={2025},
  volume={},
  number={},
  pages={1-14},
  doi={10.1109/JBHI.2025.3622725}}

@INPROCEEDINGS{Lu2024SlimSeiz,
  author={Lu, Guorui and Peng, Jing and Huang, Bingyuan and Gao, Chang and Stefanov, Todor and Hao, Yong and Chen, Qinyu},
  booktitle={2025 IEEE International Symposium on Circuits and Systems (ISCAS)}, 
  title={SlimSeiz: Efficient Channel-Adaptive Seizure Prediction Using a Mamba-Enhanced Network}, 
  year={2025},
  volume={},
  number={},
  pages={1-5},
  doi={10.1109/ISCAS56072.2025.11043364}}

@ARTICLE{Chen2025SSTDA,
  author={Chen, Peiyin and Liu, Xiaofeng and Ma, Chao and Wang, He and Yang, Xiong and Grebogi, Celso and Gu, Xiao and Gao, Zhongke},
  journal={IEEE Journal of Biomedical and Health Informatics}, 
  title={Unsupervised Domain Adaptation With Synchronized Self-Training for Cross- Domain Motor Imagery Recognition}, 
  year={2025},
  volume={29},
  number={5},
  pages={3664-3677},
  doi={10.1109/JBHI.2025.3525577}}

@article{Liu2026SSAS,
title = {SSAS: Cross-subject EEG-based emotion recognition through source selection with adversarial strategy},
journal = {Expert Systems with Applications},
volume = {308},
pages = {130843},
year = {2026},
issn = {0957-4174},
doi = {https://doi.org/10.1016/j.eswa.2025.130843},
url = {https://www.sciencedirect.com/science/article/pii/S0957417425044586},
author = {Yici Liu and Qi Wei Oung and Hoi Leong Lee}
}

@ARTICLE{Shen2025UADAAN,
  author={Shen, Jian and You, Lechun and Ma, Yu and Zhao, Zeguang and Liang, Huajian and Zhang, Yanan and Hu, Bin},
  journal={IEEE Transactions on Affective Computing}, 
  title={UA-DAAN: An Uncertainty-Aware Dynamic Adversarial Adaptation Network for EEG-Based Depression Recognition}, 
  year={2025},
  volume={16},
  number={3},
  pages={2130-2141},
  doi={10.1109/TAFFC.2025.3555433}}

@article{
Kaiser2012,
author = {Vera Kaiser  and Ian Daly  and Floriana Pichiorri  and Donatella Mattia  and Gernot R. Müller-Putz  and Christa Neuper },
title = {Relationship Between Electrical Brain Responses to Motor Imagery and Motor Impairment in Stroke},
journal = {Stroke},
volume = {43},
number = {10},
pages = {2735-2740},
year = {2012},
doi = {10.1161/STROKEAHA.112.665489},
URL = {https://www.ahajournals.org/doi/abs/10.1161/STROKEAHA.112.665489},
eprint = {https://www.ahajournals.org/doi/pdf/10.1161/STROKEAHA.112.665489}}

@article{Tangwiriyasakul2014,
doi = {10.1088/1741-2560/11/3/036001},
url = {https://doi.org/10.1088/1741-2560/11/3/036001},
year = {2014},
month = {apr},
publisher = {IOP Publishing},
volume = {11},
number = {3},
pages = {036001},
author = {Tangwiriyasakul, Chayanin and Mocioiu, Victor and van Putten, Michel J A M and Rutten, Wim L C},
title = {Classification of motor imagery performance in acute stroke},
journal = {Journal of Neural Engineering}
}

@Article{StrokeSMRReview2023,
author={Kancheva, Ivana
and van der Salm, Sandra M. A.
and Ramsey, Nick F.
and Vansteensel, Mariska J.},
title={Association between lesion location and sensorimotor rhythms in stroke -- a systematic review with narrative synthesis},
journal={Neurological Sciences},
year={2023},
month={Dec},
day={01},
volume={44},
number={12},
pages={4263-4289},
issn={1590-3478},
doi={10.1007/s10072-023-06982-8},
url={https://doi.org/10.1007/s10072-023-06982-8}
}

@Article{Liu2024XWStroke,
author={Liu, Haijie
and Wei, Penghu
and Wang, Haochong
and Lv, Xiaodong
and Duan, Wei
and Li, Meijie
and Zhao, Yan
and Wang, Qingmei
and Chen, Xinyuan
and Shi, Gaige
and Han, Bo
and Hao, Junwei},
title={An EEG motor imagery dataset for brain computer interface in acute stroke patients},
journal={Scientific Data},
year={2024},
month={Jan},
day={25},
volume={11},
number={1},
pages={131},
issn={2052-4463},
doi={10.1038/s41597-023-02787-8},
url={https://doi.org/10.1038/s41597-023-02787-8}
}

@misc{Jia2019,
author={Tianyu Jia},
title={EEG data of motor imagery for stroke},
year={2019},
month={2},
url={https://figshare.com/articles/dataset/EEG_data_of_motor_imagery_for_stroke_patients/7636301},
doi = {10.6084/m9.figshare.7636301.v6}
}

@article{Maaten2008,
  author  = {Laurens van der Maaten and Geoffrey Hinton},
  title   = {Visualizing Data using t-SNE},
  journal = {Journal of Machine Learning Research},
  year    = {2008},
  volume  = {9},
  number  = {86},
  pages   = {2579--2605},
  url     = {http://jmlr.org/papers/v9/vandermaaten08a.html}
}

@article{Kancheva2023Association,title={Association between lesion location and sensorimotor rhythms in stroke – a systematic review with narrative synthesis},author={Ivana Kancheva and Sandra M. A. van der Salm and N. Ramsey and M. Vansteensel},journal={Neurological Sciences},year={2023},volume={44},pages={4263 - 4289},doi={10.1007/s10072-023-06982-8}}

@article{Xu2022Time-Varying,title={Time-Varying Effective Connectivity for Describing the Dynamic Brain Networks of Post-stroke Rehabilitation},author={Fangzhou Xu and Yuandong Wang and Han Li and Xin Yu and Chongfeng Wang and Ming Liu and Lin Jiang and Chao Feng and Jianfei Li and De-quan Wang and Zhiguo Yan and Yang Zhang and Jiancai Leng},journal={Frontiers in Aging Neuroscience},year={2022},volume={14},doi={10.3389/fnagi.2022.911513}}

@article{Rustamov2022,title={Theta–gamma coupling as a cortical biomarker of brain–computer interface-mediated motor recovery in chronic stroke},author={N. Rustamov and Joseph B. Humphries and A. Carter and E. Leuthardt},journal={Brain Communications},year={2022},volume={4},doi={10.1093/braincomms/fcac136}}

@ARTICLE{11309705,
  author={Lu, Rongrong and Deng, Wenchang and Gao, Tianhao and Huang, Songhua and Zhang, Zhi and Liu, Yan and Zhong, Sheng-hua},
  journal={IEEE Journal of Biomedical and Health Informatics}, 
  title={Mutual Generation for Cross-Domain Challenge in Stroke Patients' Motor Imagery Classification and Functional Recovery Prediction}, 
  year={2025},
  volume={},
  number={},
  pages={1-14},
  doi={10.1109/JBHI.2025.3646871}}

@INPROCEEDINGS{11429349,
  author={Mim, Milakul Hasana and Pranto, Debjoty Roy and Hasan, Md Mehedi},
  booktitle={2026 5th International Conference on Electrical, Computer \& Telecommunication Engineering (ICECTE)}, 
  title={FA-EEGNet: Enhancing EEGNet with Frequency-Adaptive Kernels for the classification of stroke rehabilitation movements}, 
  year={2026},
  volume={},
  number={},
  pages={1-6},
  doi={10.1109/ICECTE69292.2026.11429349}}

@ARTICLE{11328767,
  author={Raza, Aaqib and Yusoff, Mohd Zuki},
  journal={IEEE Sensors Letters}, 
  title={An MAML-Based Lightweight Neural Network With Domain Adaptation for Cross-Subject and Generalized EEG-Based Motor Imagery Classification}, 
  year={2026},
  volume={10},
  number={2},
  pages={1-4},
  doi={10.1109/LSENS.2026.3650795}}

@misc{he2026,
      title={Toward Robust, Reproducible, and Widely Accessible Intracranial Language Brain-Computer Interfaces: A Comprehensive Review of Neural Mechanisms, Hardware, Algorithms, Evaluation, Clinical Pathways and Future Directions}, 
      author={Dongyi He and Wai Ting Siok and Nizhuan Wang},
      year={2026},
      eprint={2603.12279},
      archivePrefix={arXiv},
      primaryClass={q-bio.NC},
      url={https://arxiv.org/abs/2603.12279}, 
}

@ARTICLE{wang2025,
  author={Li, Yueyang and Zeng, Weiming and Dong, Wenhao and Han, Di and Chen, Lei and Chen, Hongyu and Kang, Zijian and Gong, Shengyu and Yan, Hongjie and Siok, Wai Ting and Wang, Nizhuan},
  journal={IEEE Transactions on Instrumentation and Measurement}, 
  title={A Tale of Single-Channel Electroencephalography: Devices, Datasets, Signal Processing, Applications, and Future Directions}, 
  year={2025},
  volume={74},
  number={},
  pages={1-20},
  doi={10.1109/TIM.2025.3556900}}

@article{Keser2022Electroencephalogram,title={Electroencephalogram (EEG) With or Without Transcranial Magnetic Stimulation (TMS) as Biomarkers for Post-stroke Recovery: A Narrative Review},author={Zafer Keser and Samuel C. Buchl and Nathan A. Seven and M. Markota and H. Clark and David T. Jones and G. Lanzino and Robert D. Brown and G. Worrell and B. Lundstrom},journal={Frontiers in Neurology},year={2022},volume={13},doi={10.3389/fneur.2022.827866}}

@ARTICLE{10.3389/fneur.2026.1672882,
AUTHOR={Lin, Yuhuang  and Yuan, Yong  and Chen, Jingjing  and Lin, Xiangfu },
TITLE={Motor imagery combined with brain-computer interface for stroke patients: a meta-analysis},
JOURNAL={Frontiers in Neurology},
VOLUME={Volume 17 - 2026},
YEAR={2026},
URL={https://www.frontiersin.org/journals/neurology/articles/10.3389/fneur.2026.1672882},
DOI={10.3389/fneur.2026.1672882},
ISSN={1664-2295}}

@ARTICLE{11345330,
  author={Wang, Junlin and Li, Xiaodong and Lu, Xingyu and Fei, Ningbo and Huang, Wei and Hu, Yong},
  journal={IEEE Transactions on Instrumentation and Measurement}, 
  title={A Somatosensory ERP-BCI System for Post-Stroke Hand Function Rehabilitation Applications}, 
  year={2026},
  volume={75},
  number={},
  pages={1-12},
  doi={10.1109/TIM.2026.3652757}}

\end{document}